\documentclass[letterpaper, 10 pt, conference]{ieeeconf}  

\IEEEoverridecommandlockouts                              

\usepackage{balance} 
\usepackage{amssymb}
\usepackage{graphics}
\usepackage{graphicx}
\usepackage{amsmath} 
\usepackage{floatrow}
\usepackage{url}
\usepackage{svg}
\usepackage{cite}
\usepackage{textcomp}
\usepackage{gensymb}
\usepackage[ruled,vlined]{algorithm2e}
\usepackage{multirow}
\usepackage{tablefootnote}
\usepackage{siunitx}
\usepackage{tabularray}

\usepackage{multicol}

\title{\LARGE \bf
An Approach to Combining Video and Speech with Large Language Models in Human-Robot Interaction
}
\author{Shen Guanting, Tian Zi
\thanks{The authors are with the Dalian University of Technology. Dalian, China. {\tt\footnotesize \{shen.guanting.99, zi.tian.robotics\}@gmail.com}. The first one is the corresponding author.}%
}

\begin{document}

\maketitle
\thispagestyle{empty}
\pagestyle{empty}

\begin{abstract}

Interpreting human intent accurately is a central challenge in human–robot interaction (HRI) and a key requirement for achieving more natural and intuitive collaboration between humans and machines. This work presents a novel multimodal HRI framework that combines advanced vision–language models, speech processing, and fuzzy logic to enable precise and adaptive control of a Dobot Magician robotic arm. The proposed system integrates Florence-2 for object detection, Llama 3.1 for natural language understanding, and Whisper for speech recognition, providing users with a seamless and intuitive interface for object manipulation through spoken commands. By jointly addressing scene perception and action planning, the approach enhances the reliability of command interpretation and execution. Experimental evaluations conducted on consumer-grade hardware demonstrate a command execution accuracy of 75\%, highlighting both the robustness and adaptability of the system. Beyond its current performance, the proposed architecture serves as a flexible and extensible foundation for future HRI research, offering a practical pathway toward more sophisticated and natural human–robot collaboration through tightly coupled speech and vision–language processing.

\end{abstract}

\begin{keywords}

Object Detection, Speech Recognition, LLMs, Fuzzy Logic

\end{keywords}

\section{Introduction}\label{sec:introduction}

Human–robot interaction (HRI) has rapidly evolved into a core research area as robots are increasingly deployed across diverse application domains, including healthcare, education, industry, and domestic environments~\cite{hu2011advanced, zhao2022human, he2017educational, kagami2006home}. As robots transition from isolated automation tools to collaborative partners, their ability to correctly infer and respond to human intentions becomes a fundamental requirement. Accurate intention recognition directly impacts system efficiency, operational safety, and overall user satisfaction~\cite{huang2016anticipatory, lorentz2023pointing}. This requirement is especially critical in safety-sensitive domains such as robotic surgery, where misinterpretation of a human operator’s intent can lead to severe consequences for patients~\cite{saeidi2019autonomous}, or in cases where the human and the robot are close proximity when an error can harm the human~\cite{dominguez2023improving, dominguez2024exploring, dominguez2024force}

Designing effective HRI systems remains a challenging problem due to its inherently interdisciplinary nature. Successful interaction requires the seamless integration of computer vision, machine learning, speech and audio processing, and robotic control. To address this complexity, modern robotic systems increasingly emulate elements of human perception and cognition, enabling more intuitive and accessible interaction even for users without technical expertise.

Recent advances in artificial intelligence have significantly influenced HRI research, specially through the adoption of large language models (LLMs) like LLaMA, GPT-4 or Mistral~\cite{zu2024language, touvron2023llama, achiam2023gpt}. Within robotics, these models provide powerful mechanisms for interpreting natural language instructions and extracting actionable both implicit and explicit intent~\cite{dominguez2025human} from human speech. Building upon this, vision–language models (VLMs) extend language understanding to the visual domain by jointly reasoning over images and text. Their ability to recognize a wide range of object categories and contextual relationships has made them an increasingly attractive option for HRI applications~\cite{zhang2024interactive, gao2024physically}.

Motivated by these developments, and with the objective of taking advantage of their enormous potential in HRI~\cite{atuhurra2024leveraging}, this work proposes a multimodal human–robot interaction framework that leverages recent advances in foundation models to approximate elements of human-like communication. The system translates spoken user instructions into physical actions executed by a robotic arm equipped with a suction-based end-effector. To achieve this, the Florence-2 vision–language model~\cite{xiao2024florence} is employed for object perception, LLaMA 3.1 is used for semantic interpretation of language commands, and OpenAI’s Whisper model~\cite{radford2023robust} provides robust speech-to-text transcription. These perception and language components are tightly coupled with a fuzzy logic–based control strategy that governs the robot’s motion. This combination of inference systems and direct communication with the user tend to more acceptable to the final user~\cite{dominguez2025inference} and a way to enable proactive behaviours in the robot~\cite{dominguez2024anticipation}.

Unlike many existing HRI approaches that struggle with limited scene understanding or rigid control strategies, the proposed system integrates multimodal sensory inputs to enable more robust and context-aware decision-making. By combining vision, speech, and fuzzy logic control within a unified architecture, this work addresses key limitations of current HRI systems and demonstrates a practical pathway toward more natural, adaptive, and intuitive human–robot collaboration.

\section{System Architecture}\label{sec:architecture}

\begin{figure}[t]
	\centering
	\includegraphics[width=0.98\textwidth]{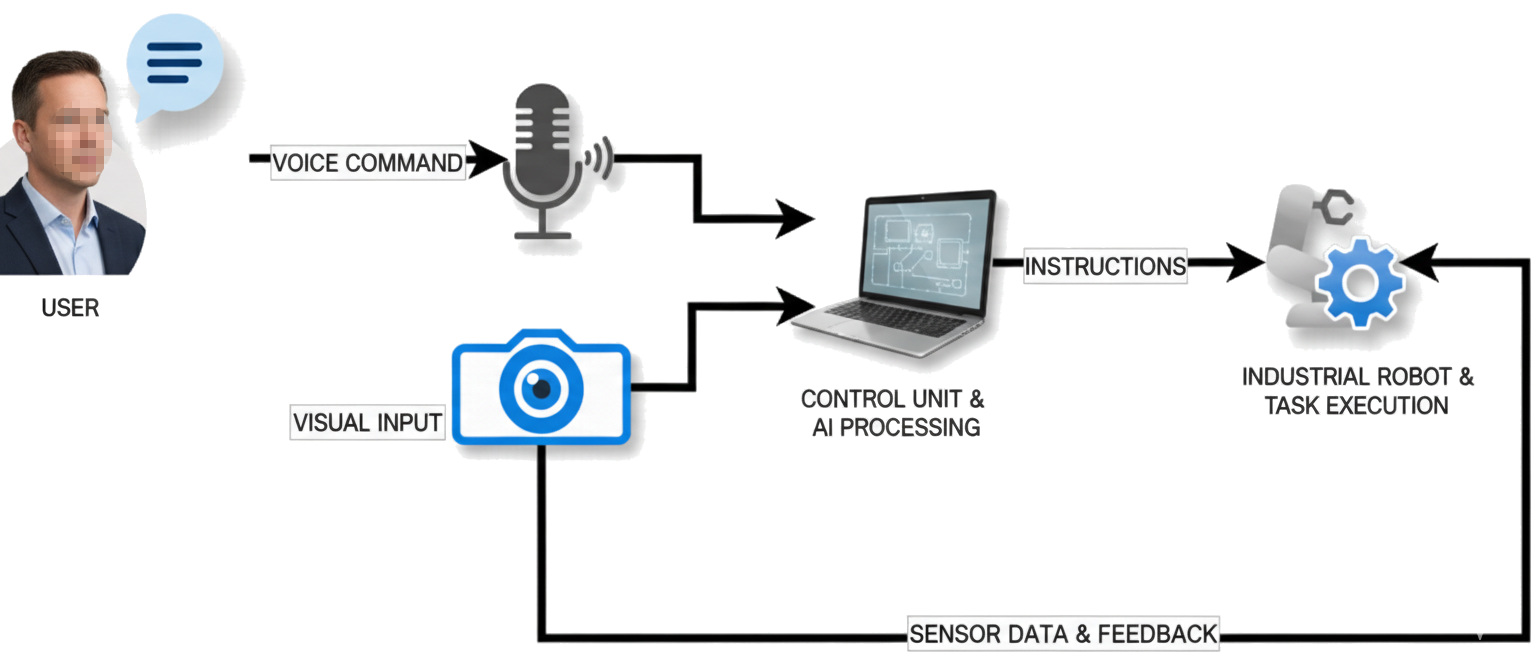}
	\caption{{\bf Information flow.} Hardware elements and data flow of the robotic manipulation system.}
	\label{fig:Block-diagram}
\end{figure}

The proposed human–robot interaction framework is designed around a tightly integrated architecture that combines multimodal perception, foundation models, and intelligent control to take advantage of the potentials of each model and technique~\cite{atuhurra2024leveraging} to enable reliable and intuitive robotic object manipulation. The system unifies visual sensing, speech-based interaction, semantic reasoning, and adaptive motion control within a single real-time pipeline. An overview of the hardware configuration and data exchange between system components is illustrated in Fig.~\ref{fig:Block-diagram}.

\subsection{Hardware Components}

The experimental setup is centered on a Dobot Magician robotic arm, which serves as the physical manipulation platform. Visual perception is provided by an Intel RealSense D435i RGB-D camera, enabling real-time scene capture for object detection and spatial awareness. Computation is performed on a high-performance laptop equipped with an Intel Core i9-14900HX processor and an NVIDIA RTX 4070 GPU, allowing concurrent execution of multiple AI models. Audio input is captured using Samsung Buds2 wireless earbuds, which provide sufficient quality for continuous speech recognition and wake-word detection.

\subsection{Processing Distribution}

To ensure real-time performance while accommodating hardware constraints, computational tasks are distributed across available CPU and GPU resources. The LLaMA 3.1 language model is executed on the GPU to accelerate natural language understanding and reduce inference latency. In contrast, Florence-2, Whisper, and the Audio Spectrogram Transformer (AST) operate on the CPU. In addition to object detection, Florence-2 is also responsible for processing camera images used by the graphical user interface and for assisting with ArUco marker–based localization of the robot’s end-effector.

User interaction begins through spoken commands, which are continuously monitored by the audio processing pipeline. Speech signals are captured and first analyzed for wake-word detection. Once activated, OpenAI’s Whisper model converts the spoken input into textual form. The resulting transcription is then forwarded to the language understanding module, where it is interpreted to determine the intended action and relevant target objects. This audio-driven interaction mechanism enables hands-free and natural communication between the user and the robotic system.

Robotic motion is governed by a fuzzy logic–based control framework that incorporates both Type-1 and Interval Type-2 Fuzzy Logic Systems (IT2FLS). These controllers translate high-level action commands into smooth and precise movements of the Dobot Magician arm. The use of interval Type-2 fuzzy logic enhances robustness by explicitly modeling uncertainty, resulting in more stable behavior during object approach and manipulation tasks.

\subsection{System and Data Workflow}

The system operates as a sequential yet tightly coupled pipeline. Interaction begins when the user issues a voice command, triggering simultaneous audio processing and visual scene capture. The camera stream is analyzed to detect ArUco markers for end-effector localization, while Florence-2 identifies objects of interest within the scene. The transcribed speech is processed by LLaMA 3.1, which extracts the intended action and determines the relevant object relationships. Based on this information, the fuzzy logic controller computes the required motion commands, which are then executed by the robotic arm. Throughout execution, continuous visual feedback ensures accurate tracking, while system status and detection results are presented through the graphical user interface.

\subsection{User Interaction}

The system is designed to prioritize intuitive and transparent user interaction. Spoken language serves as the primary input modality for selecting objects and issuing commands. Visual feedback is provided through a graphical interface that displays live camera feeds, detected objects, and ArUco marker positions in real time. This combination of auditory and visual feedback allows users to easily monitor system behavior and understand the robot’s decision-making process.

The key contribution of this architecture lies in its unified integration of foundation models for perception and language understanding with interval Type-2 fuzzy logic control for adaptive manipulation. By consolidating these components on a portable, high-performance computing platform, the system achieves real-time operation while maintaining flexibility for deployment and future extension.

\section{Methods}

\subsection{Robot End-Effector Detection}

Accurate localization of the robot’s end-effector is a fundamental requirement for reliable manipulation, as it serves as a primary reference point for motion planning and control. In the proposed system, the end-effector position is continuously estimated to ensure precise alignment between perceived object locations and executed robotic actions. To achieve robust tracking, an ArUco fiducial marker was rigidly attached to the Dobot Magician’s arm, enabling consistent detection using the OpenCV library.

\begin{figure}[t]
	\centering
	\includegraphics[width=0.98\textwidth]{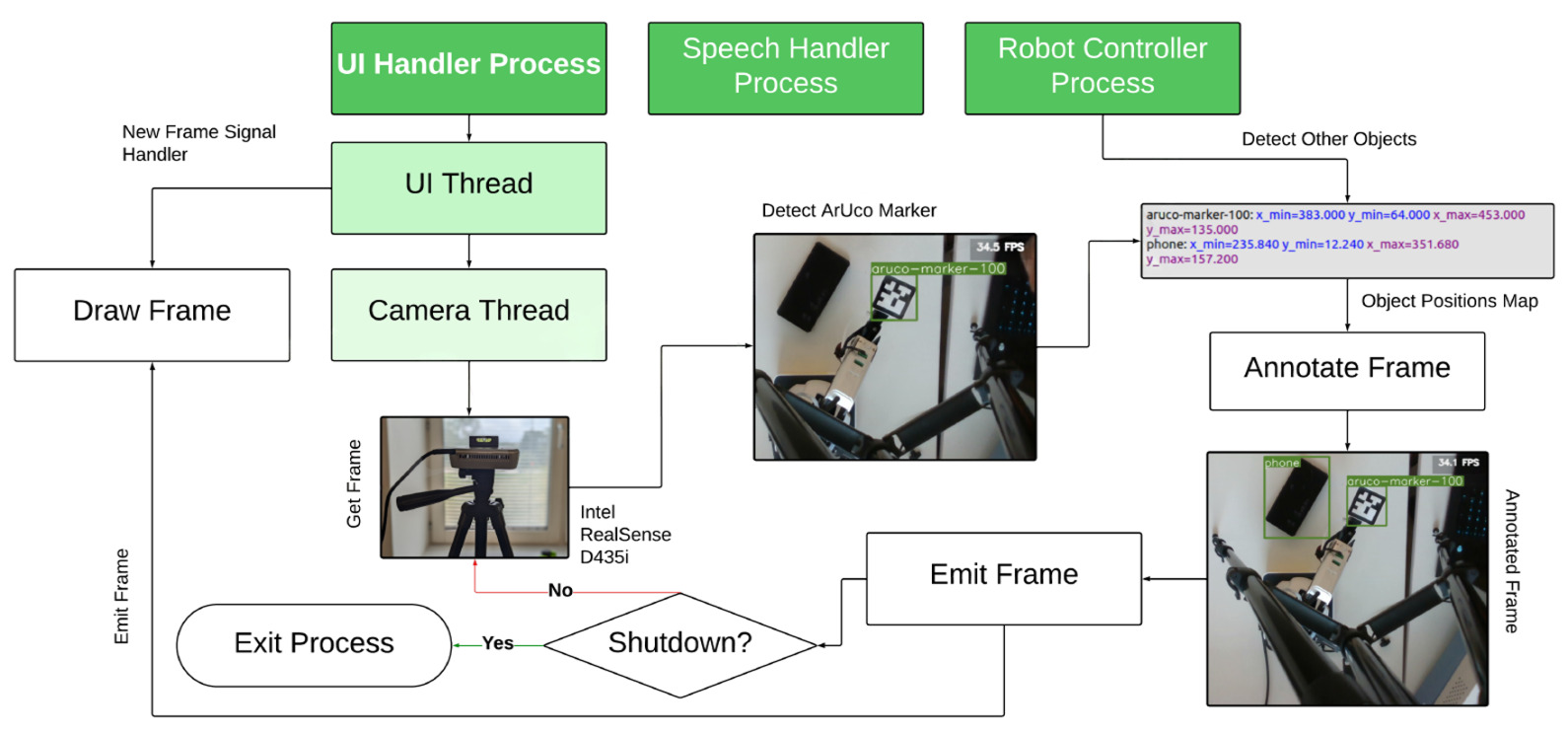}
	\caption{{\bf UI Handler Process.}}
	\label{fig:ui-process}
\end{figure}

A marker with a physical size of 100 units, selected from the 6×6 ArUco dictionary, was employed to balance detection reliability and computational efficiency. The detection process is executed within the Camera Thread of the system architecture, as illustrated in Fig.~\ref{fig:ui-process}, allowing end-effector localization to be updated in real time alongside scene perception.

Prior to adopting the marker-based approach, an alternative solution based on learning-driven detection was investigated. Specifically, a YOLOv10 object detection model~\cite{wang2024yolov10} was trained to identify the robot’s end-effector directly from camera images. While offline training and validation metrics suggested strong performance, the model failed to generalize reliably during live deployment. Despite high apparent accuracy during training, real-world inference produced inconsistent and inaccurate detections.

This discrepancy is attributed to the limited visual distinctiveness of the end-effector itself. From the camera’s perspective, the end-effector appears as a simple, predominantly black rectangular shape, offering insufficient discriminative features for robust detection across varying lighting conditions and viewpoints. As a result, the learning-based approach proved unsuitable for dependable real-time localization.

In contrast, the ArUco marker solution demonstrated superior performance in both accuracy and speed. The marker could be consistently detected at approximately 30 frames per second, providing stable and low-latency estimates of the end-effector position. This reliability significantly simplifies downstream control, as the system maintains continuous access to up-to-date end-effector coordinates at every frame, enabling precise and responsive robotic motion.

\subsection{Speech Processing}

Speech-based interaction forms the primary communication channel between the user and the robotic system. To support natural and efficient dialogue while maintaining real-time performance, the speech processing pipeline is organized into three sequential stages: wake-up command detection, speech-to-text transcription, and semantic action extraction. This modular design allows the system to remain responsive while minimizing unnecessary computational overhead.

\subsubsection{Wake-Up Command Recognition}

Continuous wake-up command detection is a critical component of the speech processing pipeline, as it enables hands-free system activation without requiring constant high-cost speech transcription. The system continuously monitors incoming audio streams for a predefined activation phrase before transitioning into full command processing mode.

For this task, a pre-trained Audio Spectrogram Transformer (AST) model—specifically the MIT/ast-finetuned-speech-commands-v2 variant—was selected~\cite{gong2021ast}. This model was chosen in preference to OpenAI’s Whisper due to its significantly smaller parameter count (approximately 85 million compared to Whisper’s 244 million) and its specialization in short command recognition. These characteristics make it well suited for low-latency wake-word detection with minimal computational burden.

Unlike text-based speech models, the AST operates directly in the time–frequency domain, processing spectrogram representations of audio signals. This eliminates the need for intermediate transcription and simplifies the detection pipeline. Audio is captured in real time using a sampling frequency aligned with the classifier’s requirements. The system processes audio in overlapping segments, with a chunk duration $T_c = 2$ seconds and a streaming step size $T_s = 0.25$ seconds, resulting in an overlap of $T_o = T_c - T_s = 1.75$ seconds. This overlap ensures robust detection even when the wake-up phrase occurs near the boundary between consecutive audio windows.

Feature extraction begins with the computation of a Short-Time Fourier Transform (STFT)\cite{mitra2001digital}, expressed as:

\begin{equation}
	X(\tau, \omega) = \sum_{n} x[n]\, w[n - \tau]\, e^{-j \omega n}
\end{equation}

In addition to the STFT, complementary spectral features such as Mel-frequency cepstral coefficients (MFCCs) are extracted and provided as input to the AST model. The classifier outputs a predicted label $l$ and an associated confidence score $s$. A wake-up command is considered detected when the following condition is satisfied:

\begin{equation}
	l = l_W \quad \text{and} \quad s > \theta
\end{equation}

\noindent
where $l_W$ denotes the predefined wake-word label and $\theta$ represents a confidence threshold, set to 0.5 by default.

The system continuously processes incoming audio until either a valid wake-up command is detected or a shutdown signal is received. The use of overlapping audio segments significantly reduces the likelihood of missed detections, while adjustable parameters such as chunk size and confidence threshold allow fine-tuning of responsiveness and detection accuracy.

\subsubsection{Speech-to-Text Conversion}

Once the wake-up command has been successfully detected, the system transitions into full speech recording mode. Audio is captured continuously until a period of silence lasting five seconds is observed, as determined by the signal amplitude falling below a predefined threshold. All recorded audio segments are concatenated into a single continuous waveform to preserve contextual coherence.

This aggregated audio stream is then processed using OpenAI’s Whisper model (small variant), which performs speech-to-text transcription. The resulting textual output serves as the input to the language understanding module and forms the basis for subsequent action interpretation.

\subsubsection{Action Extraction from Text}

Following transcription, the generated text $T$ is passed to a large language model responsible for interpreting user intent and translating it into executable robot actions. In this system, LLaMA 3.1 with 7 billion parameters is employed as the language understanding backbone. This model size was selected to balance inference capability and hardware constraints, as the available NVIDIA RTX 4070 GPU with 8 GB of memory cannot support larger models in real-time operation.

The language model operates under a carefully designed system prompt $P_{sys}$, which explicitly defines the robot’s functional capabilities and operational boundaries. The prompt restricts the model from responding to general conversational queries and instead enforces a strict focus on task-oriented instruction parsing. To ensure deterministic behavior—an essential requirement for physical robot control—the model’s temperature parameter is fixed at $\tau = 0$. This configuration eliminates stochastic variability in outputs and guarantees consistent responses for identical inputs.

To further improve output reliability, the system prompt includes structured examples that demonstrate the desired input–output format. The model is instructed to return its interpretation in a machine-readable JSON-like structure, where each inferred action corresponds to a Python-style method invocation. If a command requires a single operation, the model outputs a list containing one action; if multiple steps are needed, a sequence of actions is returned. The general output structure follows the form:

\begin{equation}
	\text{actions} = [\, \text{method\_name}(\text{arg}_1, \ldots, \text{arg}_n) \,]
\end{equation}

For example, a spatial manipulation instruction may be represented as:

\begin{equation}
	[\, \text{move\_object\_to\_left\_of}(\text{apple}, \text{orange}) \,]
\end{equation}

The generated action list is subsequently parsed and transformed into a collection of key–value mappings, where each entry consists of a method name and its corresponding arguments:

\begin{equation}
	\{\text{method\_name}, [\text{arg}_1, \ldots, \text{arg}_n]\}
\end{equation}

\begin{figure}[t]
	\centering
	\includegraphics[width=0.98\textwidth]{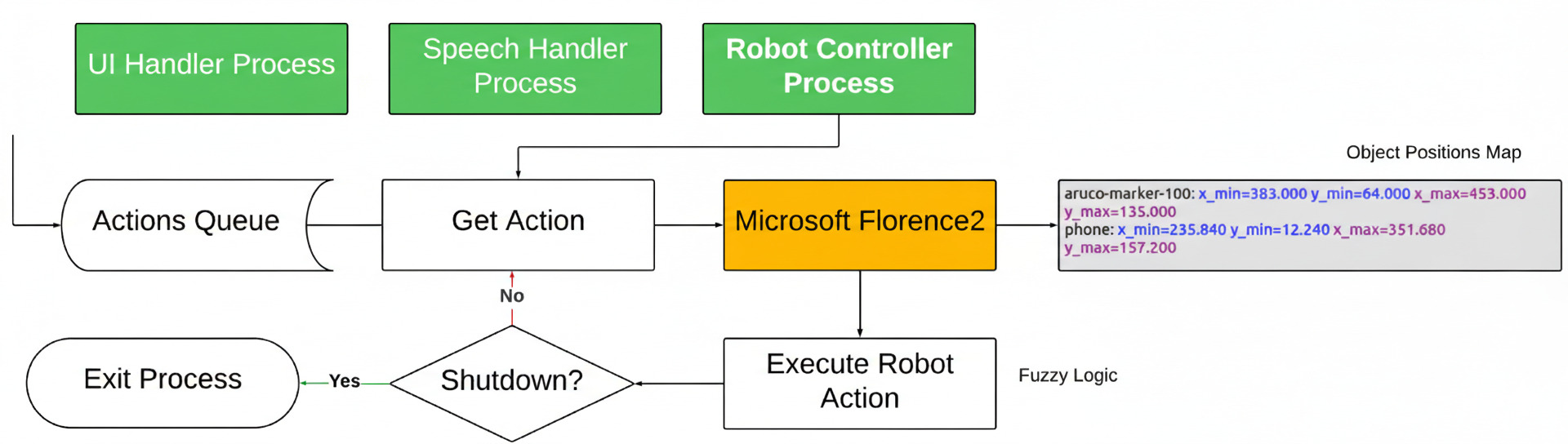}
	\caption{{\bf Robot controller.}}
	\label{fig:Robot-controller}
\end{figure}

These structured action representations are then placed into a command queue and forwarded to the robot controller, as depicted in Figure~\ref{fig:Robot-controller}. For successful execution, each referenced method must be implemented within the controller and must accept the exact number of arguments specified by the language model. This structured interface ensures reliable translation from natural language instructions to executable robotic behaviors.

\subsection{Object Detection with Florence-2}

Visual perception in the proposed system is driven by Microsoft’s Florence-2 vision foundation model, which provides flexible and prompt-driven object detection capabilities essential for natural language–based human–robot interaction. The system employs the base Florence-2 model, consisting of approximately 0.23 billion parameters, striking a balance between detection performance and computational efficiency suitable for real-time operation on consumer-grade hardware.

Florence-2 supports a wide range of visual tasks, including object detection, optical character recognition (OCR), and open-vocabulary detection. In this work, the open-vocabulary detection mode is utilized, allowing the robot to identify objects based solely on textual prompts rather than relying on a fixed, predefined set of object classes. This capability is critical for HRI scenarios, where users may refer to objects using unconstrained natural language.

Object detection is performed by issuing task-specific prompts to the model. For example, providing the prompt $<OPEN\_VOCABULARY\_DETECTION>\;green\;apple$ enables the model to locate instances of green apples within the camera’s field of view and return their corresponding bounding boxes. This prompt-based mechanism allows seamless integration between language understanding and visual perception modules.

When processing commands that involve more than one object—such as spatial relationships between items—the system generates separate detection queries for each referenced object. This design choice is necessary because Florence-2 does not support simultaneous multi-object detection within a single inference pass when operating in open-vocabulary mode. As a result, each object label is queried independently to ensure reliable localization.

Following detection, the spatial information of all recognized objects is stored in an object position map, denoted as $P_objects$. This data structure associates each detected object label with its corresponding bounding box coordinates, expressed as:

\begin{equation}
	P_{\text{objects}} = \{\text{object\_label}_i : (x_{\min}, y_{\min}, x_{\max}, y_{\max})\}
\end{equation}

\noindent
where $(x_{min}, y_{min})$ and $(x_{max}, y_{max})$ represent the top-left and bottom-right corners of the bounding box, respectively. This representation provides a compact and interpretable spatial description of object locations within the scene.

The resulting object position map serves as a key input to the robot control module. By combining object bounding box information with the estimated position of the robot’s end-effector, the fuzzy logic controller determines appropriate motion commands that guide the robotic arm toward the desired target. Through this tight coupling between open-vocabulary perception and adaptive control, the system enables flexible and context-aware object manipulation driven directly by natural language instructions.

\subsection{Fuzzy Logic Control}

Robotic manipulation in real-world environments is inherently affected by uncertainty arising from sensor noise, imprecise object localization, and actuation variability. To address these challenges and enhance the robustness of motion control, the proposed system incorporates fuzzy logic–based controllers. Both Type-1 and Interval Type-2 Fuzzy Logic Controllers (FLCs) were implemented and evaluated; however, the final system adopts an Interval Type-2 FLC due to its superior ability to explicitly model uncertainty, as established in prior studies~\cite{mendel2002fuzzy, liang2000interval}.

Unlike Type-1 fuzzy systems, which rely on fixed membership functions, Type-2 fuzzy logic introduces uncertainty directly into the membership functions themselves. This additional degree of freedom allows the controller to better accommodate imprecision in sensory inputs and environmental conditions~\cite{hellmann2001fuzzy}, making it particularly suitable for robotic manipulation tasks that require smooth and adaptive behavior.

\subsubsection{Controller Structure}

The fuzzy logic controller operates on the positional error between the robot’s end-effector and the target object. Specifically, differences along the horizontal (X) and vertical (Y) axes are used as input variables, while the corresponding corrective movements along the same axes form the controller outputs. These inputs and outputs are defined over a symmetric universe of discourse ranging from [-100, 100], discretized into 200 evenly spaced points to provide sufficient resolution for fine-grained control.

For the Interval Type-2 implementation, Gaussian membership functions were employed for both input and output fuzzy sets, following the design principles outlined in~\cite{liang2000interval}. Each variable is described using a consistent set of linguistic terms that capture varying magnitudes and directions of positional error:

\begin{itemize}
	\item NegativeVeryLarge, NegativeLarge, NegativeMedium, NegativeSmall, NegativeVerySmall
	\item Zero
	\item PositiveVerySmall, PositiveSmall, PositiveMedium, PositiveLarge, PositiveVeryLarge
\end{itemize}

This linguistic representation enables intuitive mapping between perceived positional errors and corrective actions.

\subsubsection{Rule Base}

The behavior of the controller is governed by a rule base that links input fuzzy sets to appropriate output responses. Each rule specifies how the robot should move in response to a particular combination of positional errors. For example, a negative error along the X-axis—indicating that the end-effector is positioned to one side of the target—results in a positive corrective movement to reduce the discrepancy.

The rule base was initially designed for a Type-1 FLC and subsequently extended to the Interval Type-2 framework. In the Type-2 configuration, each rule evaluates ranges of membership values rather than single-point degrees, allowing uncertainty to propagate through the inference process. Comparative testing of both controllers showed that while the Type-1 FLC provided acceptable performance, the Type-2 FLC yielded smoother motion trajectories and more stable behavior, consistent with observations reported in similar robotic applications~\cite{hagras2004type}.

\subsubsection{Optimization and Defuzzification}

To further improve control efficiency and precision, several optimization mechanisms were incorporated into the controller design. A variable step-size strategy adjusts the magnitude of movement commands based on the robot’s distance from the target, enabling larger corrections when the end-effector is far away and progressively finer adjustments as it approaches the desired position. Additionally, a dead-zone region was introduced to suppress unnecessary micro-corrections when the positional error falls below a predefined threshold, reducing oscillations and actuator wear.

For fuzzy inference, the minimum t-norm was used to model set intersections, while the maximum t-conorm was applied for set unions, in accordance with standard fuzzy control practices~\cite{mendel2002fuzzy}. Defuzzification is performed using the center-of-sets method, which offers an effective compromise between computational efficiency and output accuracy and is well suited for real-time control applications~\cite{liang2000interval}.

\section{Experiments Setup and Results}

To assess the effectiveness and practical viability of the proposed multimodal human–robot interaction framework, a comprehensive experimental evaluation was conducted using the hardware configuration described earlier. The experiments were designed to measure both the accuracy and temporal performance of the system across the full interaction pipeline, from speech input to physical task execution.

\subsection{Experimental Setup and Task Description}

A total of 60 experimental trials were performed. In each trial, the robotic system was required to interpret a spoken user command, identify the referenced object within the scene, and execute the corresponding manipulation task. Due to physical constraints of the Dobot Magician’s suction-based end-effector, real objects were represented using printed photographs of fruits placed on a flat surface. This setup ensured consistent object appearance while allowing controlled evaluation of perception and manipulation capabilities.

The task sequence involved the robot listening for a wake-up command, transcribing the user’s speech, extracting the intended action, detecting the target object(s), and executing the appropriate motion. The evaluation process recorded performance at multiple stages of the pipeline to isolate sources of delay and error. Figure 5 illustrates the experimental environment and detected objects during task execution.

\subsection{Evaluation Metrics}

System performance was analyzed using both time-based and accuracy-based metrics, computed independently for each major processing stage:

\begin{itemize}
	\item Speech-to-Text (STT): transcription time $T_{STT}$ and transcription accuracy $A_{STT}$.
	\item Action Extraction (AE): language interpretation time $T_{AE}$ and accuracy $A_{AE}$.
	\item Object Detection (OD): detection time $T_{OD}$ and accuracy $A_{OD}$.
	\item Robot Actions (RA): execution time $T_{RA}$ and accuracy $A_{RA}$.
\end{itemize}

In addition to these individual metrics, an aggregated measure was computed to capture overall system performance. The total task completion time $T_{total}$ was defined as:

\begin{equation}
	T_{\text{total}} = T_{\text{STT}} + T_{\text{AE}} + T_{\text{OD}} + T_{\text{RA}} + C
\end{equation}

\noindent
where $C$ accounts for auxiliary delays such as system overhead and user input handling. The corresponding aggregated accuracy $A_{total}$ reflects whether the system successfully completed the requested task end-to-end. Accuracy was treated as a binary outcome—either 100\% for correct execution or 0\% otherwise. Timing measurements were taken from the moment the wake-up word was detected until the robot completed the commanded action.

\subsection{Quantitative Results}

A summary of the experimental results across all 60 trials is provided in Table~\ref{tab:performance_metrics}, which reports the mean, standard deviation, and observed range for each metric.

\begin{table}[t]
	\centering
	\caption{Summary of performance metrics across 60 test cases.}
	\label{tab:performance_metrics}
	\begin{tabular}{lccc}
		\hline
		\textbf{Metric} & \textbf{Mean} & \textbf{Std Dev. (SD)} & \textbf{Range} \\
		\hline
		$T_{\text{STT}}$ (s) & 3.41  & 0.76  & 2.90--6.45 \\
		$T_{\text{AE}}$ (s)  & 3.40  & 0.75  & 2.27--6.27 \\
		$T_{\text{OD}}$ (s)  & 9.09  & 1.27  & 7.13--14.39 \\
		$T_{\text{RA}}$ (s)  & 10.08 & 1.32  & 8.25--14.52 \\
		$T_{\text{total}}$ (s) & 35.37 & 2.69 & 30.36--41.53 \\
		\hline
		$A_{\text{STT}}$ (\%) & 96.67 & 18.10 & 0--100 \\
		$A_{\text{AE}}$ (\%)  & 88.33 & 32.37 & 0--100 \\
		$A_{\text{OD}}$ (\%)  & 85.00 & 36.01 & 0--100 \\
		$A_{\text{RA}}$ (\%)  & 75.00 & 43.67 & 0--100 \\
		$A_{\text{total}}$ (\%) & 75.00 & 43.67 & 0--100 \\
		\hline
	\end{tabular}
\end{table}

The results indicate that the system achieves an overall task execution accuracy $A_{total}$ of 75\%. While this demonstrates the feasibility of the proposed approach, the relatively large standard deviation of 43.67\% reveals notable variability across different trials. This inconsistency suggests that certain interaction scenarios—particularly those involving complex perception or execution steps—remain challenging for the system.

The average total task duration $T_{total}$ was measured at 35.37 seconds, with a standard deviation of 2.69 seconds. Even in the best-performing cases, task completion times exceeded 30 seconds, which is relatively high for simple manipulation commands such as “grab the apple.” These results highlight time efficiency as a key limitation for real-world deployment, particularly in applications where responsiveness is critical.

\begin{figure}[t]
	\centering
	\includegraphics[width=0.98\textwidth]{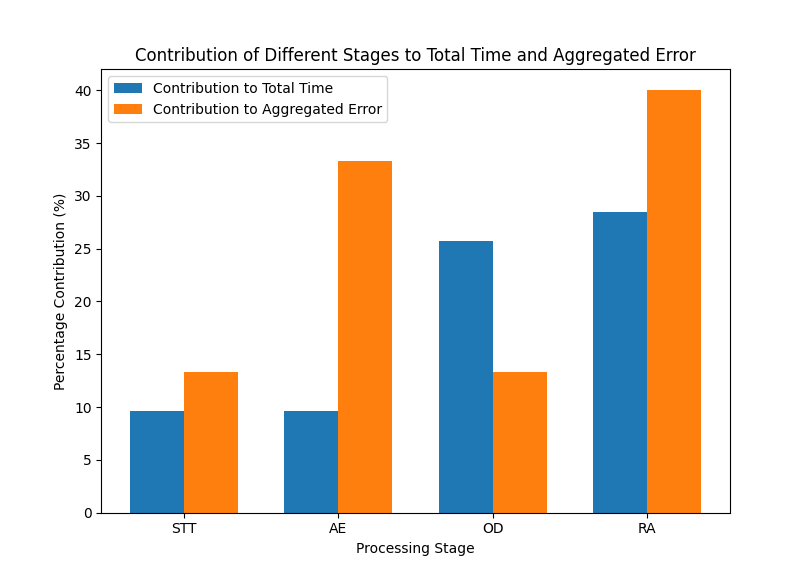}
	\caption{{\bf Percentage contribution of different stages to total time taken and aggregated error.}}
	\label{fig:Contribution}
\end{figure}

To better understand system bottlenecks, the contribution of each processing stage to both total execution time and aggregated error was analyzed, as illustrated in Fig.~\ref{fig:Contribution}.

From a temporal perspective, Object Detection (OD) and Robot Actions (RA) dominate overall task duration, accounting for 25.72\% and 28.49\% of the total time, respectively. This reflects the computational demands of vision-based perception and the physical constraints of robotic motion. In contrast, Speech-to-Text (STT) and Action Extraction (AE) each contribute less than 10\% to total execution time, demonstrating that language processing components operate relatively efficiently.

The error contribution analysis reveals a different trend. Robot Actions (RA) are responsible for the largest proportion of failures, contributing 40\% of the aggregated error. Action Extraction (AE) follows closely at 33.33\%, indicating that misinterpretation of user intent remains a significant challenge despite deterministic language model configuration. Object Detection (OD) and Speech-to-Text (STT) each contribute 13.33\% to the total error, suggesting that while object detection is time-intensive, it is comparatively reliable in terms of accuracy.

\begin{figure*}[t]
	\centering
	\begin{tabular}{cc}
		\includegraphics[width=0.460\textwidth]{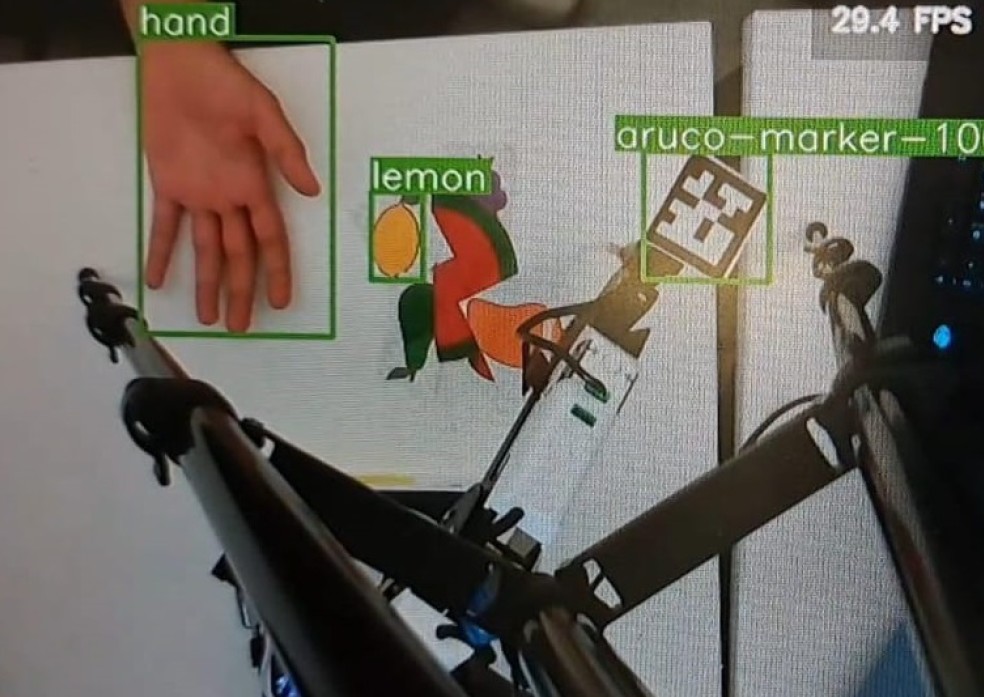} &  
		\includegraphics[width=0.460\textwidth]{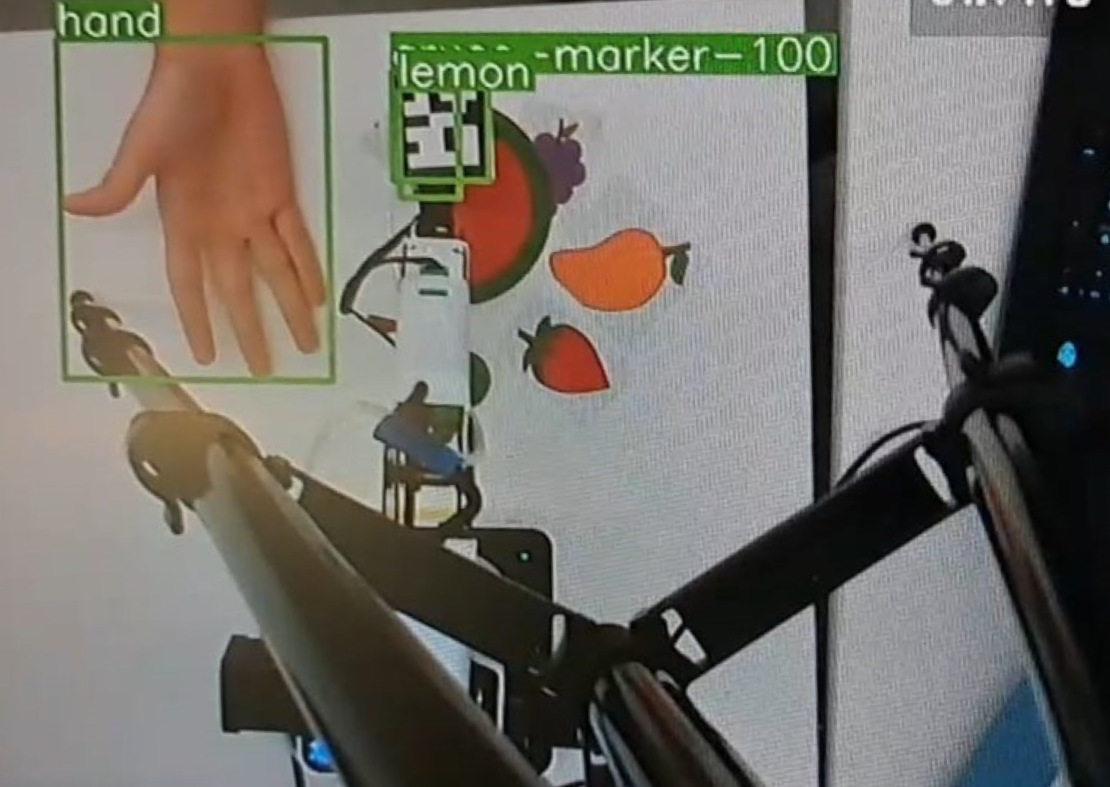} \\
	\end{tabular}
	\caption{{\bf Lemon and hand detection.} {\it Left}~-~The system identifies the user’s hand and the lemon object. {\it Right}~-~The robot positions itself to pick up the selected object.}
	\label{fig:Examples}
\end{figure*}

Some representative interaction scenarios are shown in Fig.~\ref{fig:Examples} to illustrate system behavior in practice. In Fig.~\ref{fig:Examples}~-~Left, the system successfully identifies both a lemon and the user’s hand, marking each with bounding boxes to establish spatial context. Fig.~\ref{fig:Examples}~-~Right demonstrates the robot positioning itself above the selected object and performing a pick-up action in preparation for handover.

Overall, the experimental results confirm that the proposed multimodal architecture is capable of translating natural language instructions into meaningful robotic actions with moderate reliability. While the achieved accuracy of 75\% validates the core design, the analysis also reveals clear opportunities for improvement. In particular, reducing execution time and improving robustness in action extraction and physical manipulation remain critical areas for future optimization.

\section{Conclusions}

This work presented a multimodal human–robot interaction framework that combines visual perception, speech-based communication, advanced foundation models, and fuzzy logic control to enable intuitive robotic object manipulation. By integrating vision–language models, large language models, and adaptive control strategies within a unified system, the proposed approach demonstrates a practical pathway toward more natural and efficient collaboration between humans and robots.

Experimental evaluation across 60 interaction trials confirmed the feasibility of the architecture, achieving an overall end-to-end task success rate of 75\%. The system effectively translated spoken natural language commands into coordinated robotic actions, highlighting the potential of combining multimodal perception with semantic reasoning and fuzzy control for real-world manipulation tasks. These results represent a meaningful step forward in closing the gap between high-level human intent and low-level robotic execution.

Despite these encouraging outcomes, the experiments also revealed several limitations that must be addressed to improve system reliability and responsiveness. In particular, object detection and physical action execution emerged as the primary contributors to task completion time, while action extraction and robotic motion accounted for the majority of execution errors. These findings indicate that both perception efficiency and control robustness remain critical bottlenecks in the current implementation.

Nevertheless, the identified limitations also provide clear directions for future research. Potential improvements include optimizing visual inference pipelines, refining language-to-action mapping strategies, and further enhancing the fuzzy logic controller to reduce execution variability. Additionally, extending the framework to support more complex manipulation tasks, richer dialogue, and dynamic environments would further increase its applicability and robustness.

In summary, this study contributes to the advancement of human–robot interaction by demonstrating the viability of tightly integrating multimodal inputs and foundation models with adaptive control mechanisms. The proposed architecture and experimental insights offer a solid foundation for developing more capable, flexible, and autonomous robotic systems that can operate effectively in complex, real-world scenarios.




\bibliographystyle{IEEEtran}
\balance
\bibliography{IEEEabrv,./bib.bib}

\begin{thebibliography}{10}
\providecommand{\url}[1]{#1}
\csname url@samestyle\endcsname
\providecommand{\newblock}{\relax}
\providecommand{\bibinfo}[2]{#2}
\providecommand{\BIBentrySTDinterwordspacing}{\spaceskip=0pt\relax}
\providecommand{\BIBentryALTinterwordstretchfactor}{4}
\providecommand{\BIBentryALTinterwordspacing}{\spaceskip=\fontdimen2\font plus
\BIBentryALTinterwordstretchfactor\fontdimen3\font minus \fontdimen4\font\relax}
\providecommand{\BIBforeignlanguage}[2]{{%
\expandafter\ifx\csname l@#1\endcsname\relax
\typeout{** WARNING: IEEEtran.bst: No hyphenation pattern has been}%
\typeout{** loaded for the language `#1'. Using the pattern for}%
\typeout{** the default language instead.}%
\else
\language=\csname l@#1\endcsname
\fi
#2}}
\providecommand{\BIBdecl}{\relax}
\BIBdecl

\bibitem{hu2011advanced}
J.~Hu, A.~Edsinger, Y.-J. Lim, N.~Donaldson, M.~Solano, A.~Solochek, and R.~Marchessault, ``An advanced medical robotic system augmenting healthcare capabilities-robotic nursing assistant,'' in \emph{2011 IEEE international conference on robotics and automation}.\hskip 1em plus 0.5em minus 0.4em\relax IEEE, 2011, pp. 6264--6269.

\bibitem{zhao2022human}
L.~Zhao, Z.~Hu, H.~Ding, S.~Ji, and J.~Yan, ``A human-robot interaction applicution based on augmented reality (ar) for industrial robot grasping process,'' in \emph{2022 7th International Conference on Robotics and Automation Engineering (ICRAE)}.\hskip 1em plus 0.5em minus 0.4em\relax IEEE, 2022, pp. 312--316.

\bibitem{he2017educational}
B.~He, M.~Xia, X.~Yu, P.~Jian, H.~Meng, and Z.~Chen, ``An educational robot system of visual question answering for preschoolers,'' in \emph{2017 2nd international conference on robotics and automation engineering (ICRAE)}.\hskip 1em plus 0.5em minus 0.4em\relax IEEE, 2017, pp. 441--445.

\bibitem{kagami2006home}
S.~Kagami, S.~Thompson, Y.~Nishida, T.~Enomoto, and T.~Matsui, ``Home robot service by ceiling ultrasonic locator and microphone array,'' in \emph{Proceedings 2006 IEEE International Conference on Robotics and Automation, 2006. ICRA 2006.}\hskip 1em plus 0.5em minus 0.4em\relax IEEE, 2006, pp. 3171--3176.

\bibitem{huang2016anticipatory}
C.-M. Huang and B.~Mutlu, ``Anticipatory robot control for efficient human-robot collaboration,'' in \emph{2016 11th ACM/IEEE International Conference on Human-Robot Interaction (HRI)}, 2016, pp. 83--90.

\bibitem{lorentz2023pointing}
V.~Lorentz, M.~Weiss, K.~Hildebrand, and I.~Boblan, ``Pointing gestures for human-robot interaction with the humanoid robot digit,'' in \emph{2023 32nd IEEE International Conference on Robot and Human Interactive Communication (RO-MAN)}.\hskip 1em plus 0.5em minus 0.4em\relax IEEE, 2023, pp. 1886--1892.

\bibitem{saeidi2019autonomous}
H.~Saeidi, H.~N. Le, J.~D. Opfermann, S.~L{\'e}onard, A.~Kim, M.~H. Hsieh, J.~U. Kang, and A.~Krieger, ``Autonomous laparoscopic robotic suturing with a novel actuated suturing tool and 3d endoscope,'' in \emph{2019 international conference on robotics and automation (ICRA)}.\hskip 1em plus 0.5em minus 0.4em\relax IEEE, 2019, pp. 1541--1547.

\bibitem{dominguez2023improving}
J.~E. Dom{\'\i}nguez-Vidal and A.~Sanfeliu, ``Improving human-robot interaction effectiveness in human-robot collaborative object transportation using force prediction,'' in \emph{2023 IEEE/RSJ International Conference on Intelligent Robots and Systems (IROS)}.\hskip 1em plus 0.5em minus 0.4em\relax IEEE, 2023, pp. 7839--7845.

\bibitem{dominguez2024exploring}
J.~E. Dominguez-Vidal and A.~Sanfeliu, ``Exploring transformers and visual transformers for force prediction in human-robot collaborative transportation tasks,'' in \emph{2024 IEEE International Conference on Robotics and Automation (ICRA)}.\hskip 1em plus 0.5em minus 0.4em\relax IEEE, 2024, pp. 3191--3197.

\bibitem{dominguez2024force}
J.~E. Dom{\'\i}nguez-Vidal and A.~Sanfeliu, ``Force and velocity prediction in human-robot collaborative transportation tasks through video retentive networks,'' in \emph{2024 IEEE/RSJ International Conference on Intelligent Robots and Systems (IROS)}.\hskip 1em plus 0.5em minus 0.4em\relax IEEE, 2024, pp. 9307--9313.

\bibitem{zu2024language}
W.~Zu, W.~Song, R.~Chen, Z.~Guo, F.~Sun, Z.~Tian, W.~Pan, and J.~Wang, ``Language and sketching: An llm-driven interactive multimodal multitask robot navigation framework,'' in \emph{2024 IEEE International Conference on Robotics and Automation (ICRA)}.\hskip 1em plus 0.5em minus 0.4em\relax IEEE, 2024, pp. 1019--1025.

\bibitem{touvron2023llama}
H.~Touvron, T.~Lavril, G.~Izacard, X.~Martinet, M.-A. Lachaux, T.~Lacroix, B.~Rozi{\`e}re, N.~Goyal, E.~Hambro, F.~Azhar \emph{et~al.}, ``Llama: Open and efficient foundation language models,'' \emph{arXiv preprint arXiv:2302.13971}, 2023.

\bibitem{achiam2023gpt}
J.~Achiam, S.~Adler, S.~Agarwal, L.~Ahmad, I.~Akkaya, F.~L. Aleman, D.~Almeida, J.~Altenschmidt, S.~Altman, S.~Anadkat \emph{et~al.}, ``Gpt-4 technical report,'' \emph{arXiv preprint arXiv:2303.08774}, 2023.

\bibitem{dominguez2025human}
J.~E. Dom{\'\i}nguez-Vidal and A.~Sanfeliu, ``The human intention: a taxonomy attempt and its applications to robotics,'' \emph{International Journal of Social Robotics}, vol.~17, no.~11, pp. 2479--2499, 2025.

\bibitem{zhang2024interactive}
Z.~Zhang, A.~Lin, C.~W. Wong, X.~Chu, Q.~Dou, and K.~S. Au, ``Interactive navigation in environments with traversable obstacles using large language and vision-language models,'' in \emph{2024 IEEE International Conference on Robotics and Automation (ICRA)}.\hskip 1em plus 0.5em minus 0.4em\relax IEEE, 2024, pp. 7867--7873.

\bibitem{gao2024physically}
J.~Gao, B.~Sarkar, F.~Xia, T.~Xiao, J.~Wu, B.~Ichter, A.~Majumdar, and D.~Sadigh, ``Physically grounded vision-language models for robotic manipulation,'' in \emph{2024 IEEE International Conference on Robotics and Automation (ICRA)}.\hskip 1em plus 0.5em minus 0.4em\relax IEEE, 2024, pp. 12\,462--12\,469.

\bibitem{atuhurra2024leveraging}
J.~Atuhurra, ``Leveraging large language models in human-robot interaction: a critical analysis of potential and pitfalls,'' \emph{arXiv preprint arXiv:2405.00693}, 2024.

\bibitem{xiao2024florence}
B.~Xiao, H.~Wu, W.~Xu, X.~Dai, H.~Hu, Y.~Lu, M.~Zeng, C.~Liu, and L.~Yuan, ``Florence-2: Advancing a unified representation for a variety of vision tasks,'' in \emph{Proceedings of the IEEE/CVF Conference on Computer Vision and Pattern Recognition}, 2024, pp. 4818--4829.

\bibitem{radford2023robust}
A.~Radford, J.~W. Kim, T.~Xu, G.~Brockman, C.~McLeavey, and I.~Sutskever, ``Robust speech recognition via large-scale weak supervision,'' in \emph{International conference on machine learning}.\hskip 1em plus 0.5em minus 0.4em\relax PMLR, 2023, pp. 28\,492--28\,518.

\bibitem{dominguez2025inference}
J.~E. Dom{\'\i}nguez-Vidal and A.~Sanfeliu, ``When the inference meets the explicitness or why multimodality can make us forget about the perfect predictor,'' \emph{International Journal of Social Robotics}, vol.~17, no.~12, pp. 2965--2980, 2025.

\bibitem{dominguez2024anticipation}
J.~E. Dominguez-Vidal and A.~Sanfeliu, ``Anticipation and proactivity. unraveling both concepts in human-robot interaction through a handover example,'' in \emph{2024 33rd IEEE International Conference on Robot and Human Interactive Communication (ROMAN)}.\hskip 1em plus 0.5em minus 0.4em\relax IEEE, 2024, pp. 957--962.

\bibitem{wang2024yolov10}
A.~Wang, H.~Chen, L.~Liu, K.~Chen, Z.~Lin, J.~Han \emph{et~al.}, ``Yolov10: Real-time end-to-end object detection,'' \emph{Advances in Neural Information Processing Systems}, vol.~37, pp. 107\,984--108\,011, 2024.

\bibitem{gong2021ast}
Y.~Gong, Y.-A. Chung, and J.~Glass, ``Ast: Audio spectrogram transformer,'' \emph{arXiv preprint arXiv:2104.01778}, 2021.

\bibitem{mitra2001digital}
S.~K. Mitra, \emph{Digital signal processing: a computer-based approach}.\hskip 1em plus 0.5em minus 0.4em\relax McGraw-Hill Higher Education, 2001.

\bibitem{mendel2002fuzzy}
J.~M. Mendel, ``Fuzzy logic systems for engineering: a tutorial,'' \emph{Proceedings of the IEEE}, vol.~83, no.~3, pp. 345--377, 2002.

\bibitem{liang2000interval}
Q.~Liang and J.~M. Mendel, ``Interval type-2 fuzzy logic systems: theory and design,'' \emph{IEEE Transactions on Fuzzy systems}, vol.~8, no.~5, pp. 535--550, 2000.

\bibitem{hellmann2001fuzzy}
M.~Hellmann, ``Fuzzy logic introduction,'' \emph{Universit{\'e} de Rennes}, vol.~1, no.~1, 2001.

\bibitem{hagras2004type}
H.~Hagras, ``A type-2 fuzzy logic controller for autonomous mobile robots,'' in \emph{2004 IEEE International conference on fuzzy systems (IEEE Cat. No. 04CH37542)}, vol.~2.\hskip 1em plus 0.5em minus 0.4em\relax IEEE, 2004, pp. 965--970.

\end{thebibliography}

\end{document}